\documentclass{article}
\usepackage[preprint]{neurips_2024}

\usepackage[utf8]{inputenc} 
\usepackage[T1]{fontenc}    
\usepackage{hyperref}       
\usepackage{url}            
\usepackage{booktabs}       
\usepackage{amsfonts}       
\usepackage{nicefrac}       
\usepackage{microtype}      
\usepackage{xcolor}         
\usepackage{graphicx}       
\usepackage{float}          

\makeatletter
\renewcommand{\@neuripsordinal}{Preprint}
\makeatother

\title{K-Dense Analyst: Towards Fully Automated Scientific Analysis}

\author{%
  Orion Li$^{1}$ \quad Vinayak Agarwal$^{1,2}$ \quad Summer Zhou$^{1}$ \quad Ashwin Gopinath$^{1,2}$ \quad Timothy Kassis$^{1}$ \\
  \\
  $^{1}$Biostate AI, Palo Alto, CA\\
  $^{2}$Bayosthiti AI, Bengaluru, India\\
  \\
  \texttt{\{firstname\}.\{lastname\}@biostate.ai}
}

\begin{document}

\maketitle

\vspace{-1em}
\begin{figure}[H]
  \centering
  \includegraphics[width=0.75\textwidth]{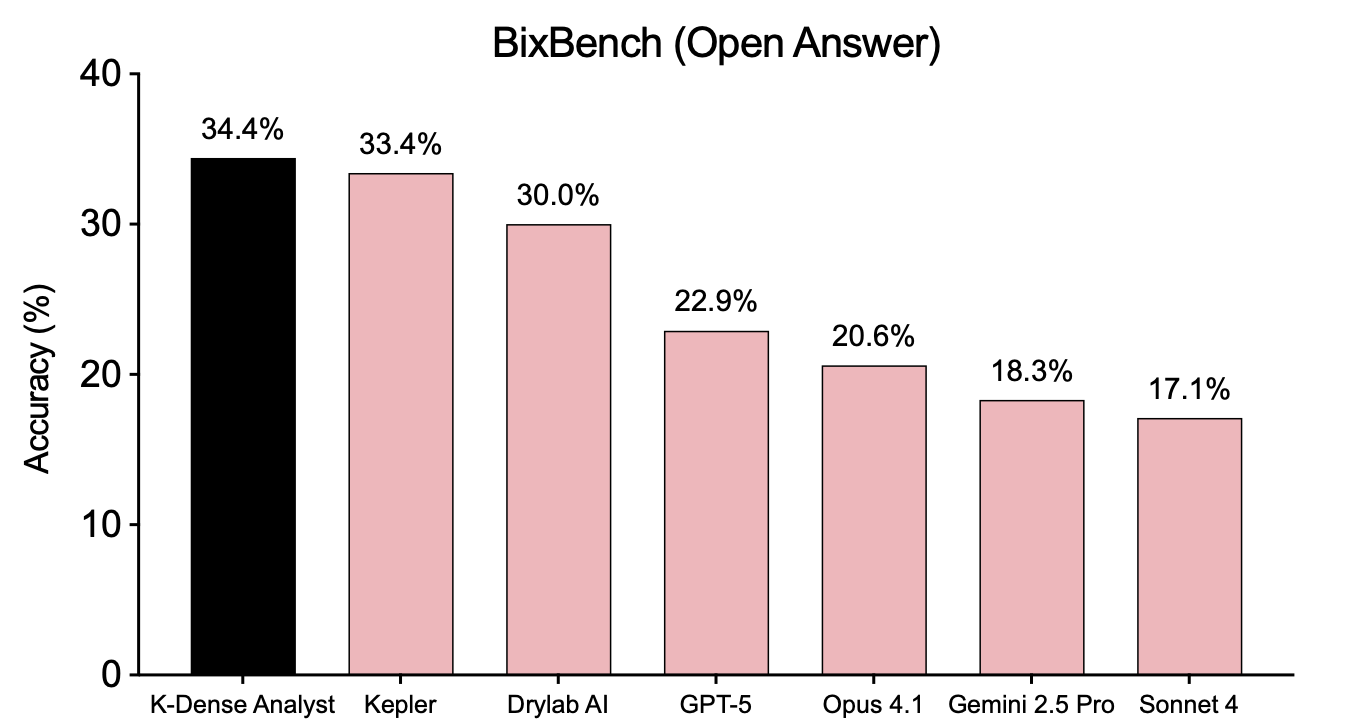}
  \caption{\textbf{K-Dense Analyst achieves state-of-the-art performance on BixBench open-answer benchmark.} Our system attains 34.4\% accuracy, surpassing GPT-5 (22.9\%) by 11.5 percentage points (while using Gemini 2.5 Pro) and the state-of-the-art agentic system from Kepler (33.4\%) by 1 percentage point.}
  \label{fig:performance-top}
\end{figure}
\vspace{-0.5em}

\begin{abstract}
The complexity of modern bioinformatics analysis has created a critical gap between data generation and developing scientific insights. While large language models (LLMs) have shown promise in scientific reasoning, they remain fundamentally limited when dealing with real-world analytical workflows that demand iterative computation, tool integration, and rigorous validation. We introduce K-Dense Analyst, a hierarchical multi-agent system that achieves autonomous bioinformatics analysis through a dual-loop architecture. K-Dense Analyst, part of the broader K-Dense platform, couples planning with validated execution using specialized agents to decompose complex objectives into executable, verifiable tasks within secure computational environments. On BixBench, a comprehensive benchmark for open-ended biological analysis, K-Dense Analyst achieves 34.4\% accuracy, surpassing the best-performing language model (GPT-5) by 11.5 percentage points over what is widely considered the most powerful LLM available. Furthermore, we surpass custom-built commercial agentic systems, such as Kepler, by at least 1\%. Our insights demonstrate that autonomous scientific reasoning requires more than enhanced language models; it demands purpose-built systems that can bridge the gap between high-level scientific objectives and low-level computational execution. These results represent a significant advance toward fully autonomous computational biologists capable of accelerating discovery across the life sciences.
\end{abstract}

\section{Introduction}

Modern biological research generates data at a pace that has outstripped human analytical capacity. A single genomics experiment can produce terabytes of information requiring dozens of specialized tools, complex statistical analyses, and deep domain expertise to interpret \cite{xiao2024cellagent, mondal2025multiagent, mitchener2025bixbench}. This analytical bottleneck has become the rate-limiting step in scientific discovery, with researchers spending months analyzing data that took days to generate.

Large language models (LLMs) have emerged as powerful tools for scientific reasoning, demonstrating remarkable capabilities in tasks ranging from literature synthesis to experimental design. Recent advances in agentic frameworks have extended these capabilities to autonomous code generation and execution, enabling AI systems to perform complex analytical workflows. However, when evaluated on real-world bioinformatics tasks, even the most advanced models struggle to achieve meaningful performance. This gap between promise and practice points to a severe limitation: current approaches remain constrained by their reliance on predefined toolsets, narrow task specifications, and the lack of robust validation mechanisms required for rigorous scientific analysis \cite{xiao2024cellagent, jin2025stella, huang2025deep, tang2025airesearcher, lu2024aiscientist}.

The challenge becomes clear when examining how bioinformatics analyses actually unfold. A typical workflow might begin with quality control of sequencing data, proceed through multiple transformation and normalization steps, incorporate statistical testing with careful multiple hypothesis correction, and conclude with biological interpretation requiring integration of pathway databases and literature knowledge. Each of these steps can fail in subtle ways, often leading to cascade failure through subsequent analyses. Without the ability to recognize these failures, adapt strategies, and validate results, no system can claim analytical autonomy.

BixBench has crystallized these challenges into a rigorous evaluation framework, revealing that frontier models achieve less than 23\% accuracy on open-ended bioinformatics questions despite their impressive performance on general reasoning benchmarks \cite{mitchener2025bixbench, zhang2025origene, yang2025mlomics}. This performance ceiling suggests that incremental improvements to existing approaches are unlikely to suffice, calling for a fundamental rethinking of the system architecture.

To address these limitations, we introduce three key contributions. First, we present K-Dense Analyst, a multi-agent framework with hierarchical planning and validation loops specifically designed for autonomous scientific analysis. Unlike existing systems that treat biological analysis as an extension of general reasoning, K-Dense Analyst recognizes it as a distinct challenge requiring specialized architectural solutions. Second, we demonstrate K-Dense Analyst's effectiveness through a comprehensive evaluation on BixBench, achieving 34.4\% accuracy that surpasses all evaluated models and agentic systems by substantial margins. Third, we provide a detailed analysis of our dual-loop architecture and its implications for building trustworthy autonomous scientific systems \cite{tang2025airesearcher, lu2024aiscientist, alber2025cellvoyager, roohani2024biodiscoveryagent}.

\section{Related Work}

The integration of large language models into scientific workflows has emerged as a transformative approach to accelerating research discovery. This work builds upon several converging research areas, including agentic AI frameworks, scientific tool automation, and specialized benchmarks for evaluating autonomous research capabilities.

\subsection{Agentic and Scaffolded LLMs}

The development of agentic AI systems represents a significant evolution from static language models to dynamic, tool-using agents capable of complex reasoning and action. ReAct frameworks have demonstrated the power of combining reasoning and acting in language models, enabling iterative problem-solving through structured thought processes \cite{yao2022react}. Recent advances in multi-agent systems have shown particular promise for scientific applications, with frameworks like AutoGPT and specialized research agents achieving remarkable capabilities in hypothesis generation and experimental design \cite{tang2025airesearcher}.

STELLA has demonstrated self-evolving capabilities in biomedical research, achieving state-of-the-art performance on challenging benchmarks through dynamic tool discovery and template learning \cite{jin2025stella}. Similarly, TxGemma and related therapeutic AI agents have shown how specialized foundation models can be fine-tuned for domain-specific scientific reasoning \cite{wang2025txgemma}.

\subsection{Scientific Tool Use and Automation}

The automation of scientific workflows through AI agents has gained significant traction across multiple domains. Systems like SciToolAgent have demonstrated the feasibility of integrating hundreds of computational tools through knowledge graph-driven approaches, enabling sophisticated multi-tool workflows \cite{ding2025scitoolagent}. In computational chemistry, ChemGraph has shown how agentic frameworks can streamline complex molecular simulations from density functional theory to machine learning potentials \cite{pham2025chemgraph}.

The Genesis project (e.g., The AI Scientist) represents an ambitious effort toward full automation of systems biology research, employing agentic methods for hypothesis-driven experimentation and manuscript generation \cite{lu2024aiscientist}. These systems collectively demonstrate the potential for AI agents to handle the complexity and diversity of modern scientific computational workflows.

\subsection{Bioinformatics Automation and Discovery}

The bioinformatics domain has seen remarkably rapid advancement in autonomous analysis systems. OLAF has pioneered conversational bioinformatics through natural language interfaces, enabling researchers to perform complex analyses without programming expertise \cite{riffle2025olaf}.

PROTEUS has demonstrated fully automated proteomics research workflows, generating novel hypotheses through hierarchical planning and specialized tool integration \cite{ding2024proteus}. The development of biomedical reasoning agents capable of handling multi-modal data has been exemplified by systems addressing single-cell analysis (e.g., CellAgent \cite{xiao2024cellagent}; CellAtria \cite{nouri2025cellatria}), genomics pipelines (e.g., CRISPR-GPT \cite{qu2024crisprgpt}), and therapeutic discovery (e.g., OriGene \cite{zhang2025origene}; BioDiscoveryAgent \cite{roohani2024biodiscoveryagent}). These advances highlight both the promise and challenges of achieving true autonomy in biological data analysis.

\subsection{Benchmarks and Evaluation Frameworks}

The assessment of autonomous scientific capabilities requires specialized benchmarks that capture the complexity of real-world research tasks. BixBench represents a landmark effort in this direction, providing 53 real-world bioinformatics scenarios with 296 open-answer questions designed to evaluate multi-step analytical reasoning \cite{mitchener2025bixbench}.

Related benchmarks such as LAB-Bench \cite{laurent2024labbench} and DiscoveryBench \cite{majumder2024discoverybench} have focused on broader biological research capabilities, including literature review and experimental planning. In the chemistry domain, ChemBench and subsequent agent analysis surveys \cite{mirza2024llmchemists} have established comprehensive evaluation frameworks for autonomous chemical discovery. The emergence of scientific agent benchmarks like ScienceAgentBench \cite{chen2024scienceagentbench} and Genome-Bench \cite{yin2025genomebench} has further expanded evaluation capabilities across diverse scientific domains. These benchmarks collectively provide the foundation for rigorous assessment of autonomous scientific discovery systems.

\section{Results}

\subsection{K-Dense Analyst System Overview}

\begin{figure}[t]
  \centering
  \includegraphics[width=\textwidth]{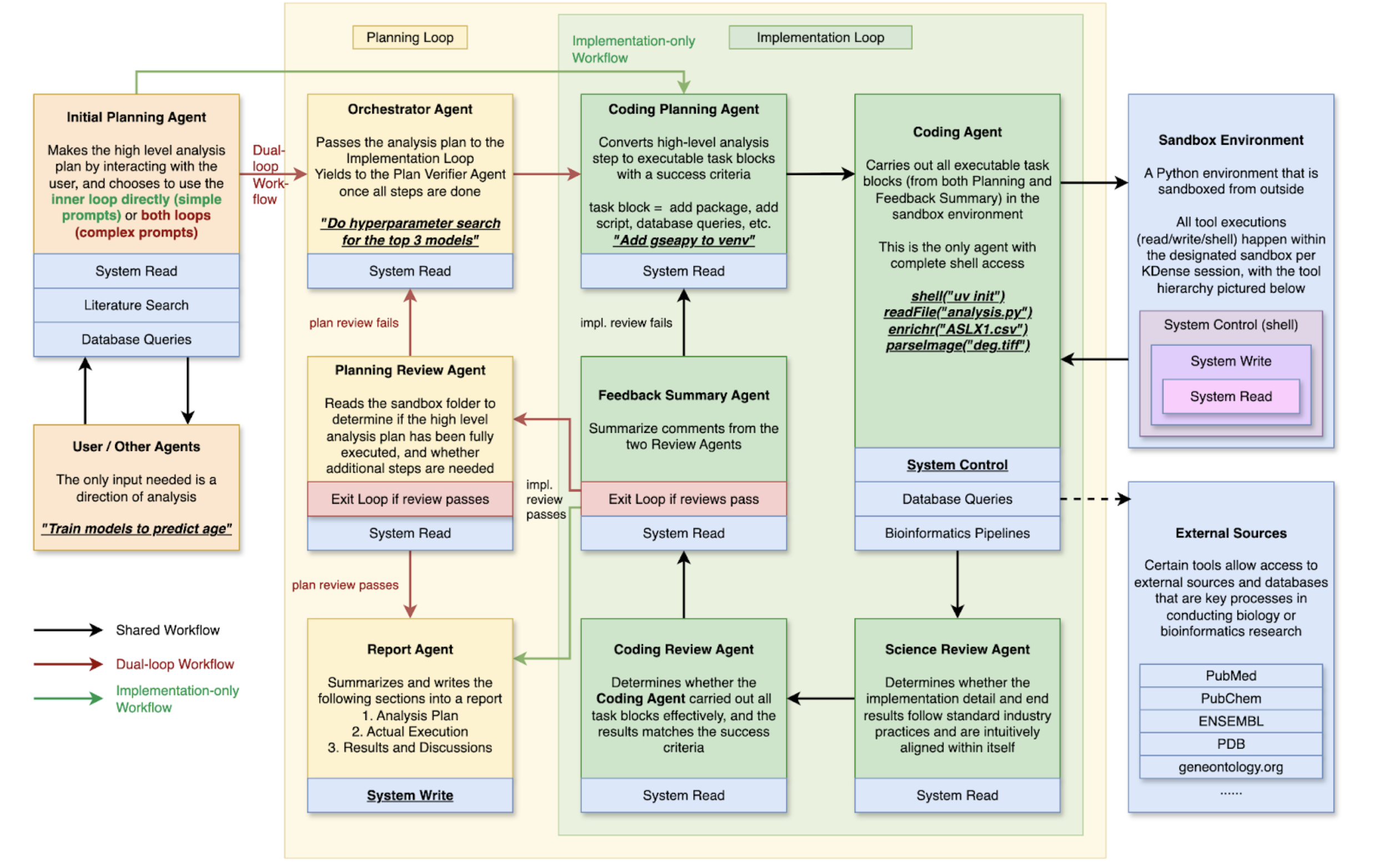}
  \caption{K-Dense Analyst architecture, showing the dual-loop workflow structure that enables both simple and complex analytical tasks. The system employs two nested feedback loops: a Planning Loop for high-level strategy development and an Implementation Loop for detailed execution and validation. This architecture allows K-Dense Analyst to handle tasks ranging from straightforward data queries to complex multi-step analyses requiring iterative refinement.}
  \label{fig:architecture}
\end{figure}

The fundamental insight driving K-Dense Analyst's design is that autonomous scientific analysis requires more than just enhanced reasoning; it demands a structured interplay between planning and execution. Our system achieves this through a dual-loop architecture (Figure \ref{fig:architecture}) that mirrors how human scientists approach complex analysis: first developing a high-level strategy, then executing detailed steps with continuous validation and refinement.

K-Dense Analyst is designed around three core principles: rigor, autonomy, and extensibility. The system aims to minimize human intervention while maintaining scientific rigor through extensive standard examples and comprehensive validation mechanisms. Unlike existing frameworks that rely on static tool libraries, K-Dense Analyst emphasizes adaptability through modular agent design and dynamic workflow generation. The framework employs a hierarchical multi-agent architecture that decomposes complex analyses into verifiable steps that autonomously iterate towards fulfilling varying scopes of different tasks. Extensibility is achieved through a modular design where new tools and analytical capabilities can be seamlessly integrated into a sub-agent without requiring system-wide modifications.

The power of the architecture lies in its dual-loop design. The planning loop operates at the strategic level, where the orchestrator agent manages multi-step analytical plans while the planning review agent ensures comprehensive coverage of scientific requirements. This loop can spawn multiple iterations, progressively building towards the success criteria articulated during initial analysis. The implementation loop handles tactical execution, with the coding planning agent decomposing objectives into executable tasks and the coding agent performing all computational work within a secure, sandboxed environment.

This separation of concerns enables K-Dense Analyst to handle both straightforward queries that can bypass planning entirely and complex multi-step analyses requiring iterative refinement. For example, the system can recognize when a simple database lookup suffices versus when a complete differential expression analysis with pathway enrichment is needed, adapting its approach accordingly.

We employ a sophisticated agent hierarchy comprising ten specialized agents, each with distinct responsibilities and capabilities. The Initial Planning Agent serves as the entry point, analyzing user requests and determining the appropriate workflow complexity. For simple tasks, the system can bypass the Planning Loop and proceed directly to implementation. For complex analyses, the Orchestrator Agent manages multi-step plan execution, coordinating with the Planning Review Agent to ensure comprehensive coverage of analytical requirements, incrementally building towards the success criteria indicated by the Initial Planning Agent.

The Implementation Loop forms the core of K-Dense Analyst's analytical capabilities. The Coding Planning Agent decomposes high-level objectives into specific executable tasks, while the Coding Agent performs all computational work within a secure sandbox environment. This agent has exclusive access to system control functions, including shell commands and file operations, ensuring security while enabling comprehensive analytical capabilities. All computational work occurs here to prevent unauthorized operations. The sandbox encloses the full spectrum of bioinformatics tools and databases required for modern analyses, as well as curated code and shell examples for an extensive list of tasks that the Coding Agent may decide to use as references. This architecture ensures that K-Dense Analyst can perform sophisticated analyses while maintaining security and reproducibility standards essential for scientific research.

Critical to the system's reliability is its multi-layer validation architecture. Within the implementation loop, the coding review agent and science review agent provide dual validation, assessing both technical implementation quality and scientific methodology validity. The feedback summary agent synthesizes review feedback and determines whether additional iterations are required. Further, the planning review agent conducts an independent review of implementation quality, suggesting key next steps for the planning loop or declaring the analysis complete. Finally, the report agent synthesizes the entire analysis trajectory into a technical report that clearly signals the achievements of the K-Dense Analyst run.

This comprehensive validation ensures that K-Dense Analyst not only executes analyses but also maintains the scientific rigor expected of human experts. The system can identify when results are inconsistent, when additional controls are needed, or when alternative analytical approaches might be more appropriate.

\subsection{Performance Results}

The true test to any autonomous scientific system lies in its ability to handle real-world analytical challenges. Figure \ref{fig:performance-top} presents K-Dense Analyst's performance on the BixBench open-answer evaluation, where the system demonstrates a decisive advantage over existing approaches.

K-Dense Analyst achieves 34.4\% accuracy on these open-answer challenges, establishing a new state-of-the-art for autonomous biological analysis. This represents an 11.5 percentage point improvement over GPT-5 (22.9\%), the best-performing language model in our evaluation. The performance gap widens further when compared to other frontier models: Opus 4.1 (20.6\%), Gemini 2.5 Pro (18.3\%), and Sonnet 4 (17.1\%). Additionally, K-Dense Analyst outperforms Kepler, a SOTA agentic system for bioinformatics, by 1\%.

These results reveal a critical insight: the challenges of scientific analysis cannot be addressed through language modeling alone. The 17.3 percentage point gap between K-Dense Analyst and Sonnet 4 represents not just incremental improvement but a fundamental difference in capability. While language models excel at pattern recognition and reasoning over text, scientific analysis demands actual computation, tool orchestration, and iterative refinement, capabilities that emerge only through purpose-built architectures.

The consistency of K-Dense Analyst's advantage across different types of bioinformatics tasks suggests that our dual-loop architecture addresses fundamental limitations in how current models approach scientific problems. Rather than attempting to solve complex analyses in a single forward pass, K-Dense Analyst can explore multiple analytical strategies, validate intermediate results, and adapt its approach based on empirical outcomes.

\subsection{Case Studies: K-Dense Analyst in Action}

To illustrate how K-Dense Analyst achieves its superior performance, we present three representative examples from BixBench that demonstrate the system's analytical approach and problem-solving capabilities.

\begin{figure}[h!]
  \centering
  \includegraphics[width=\textwidth]{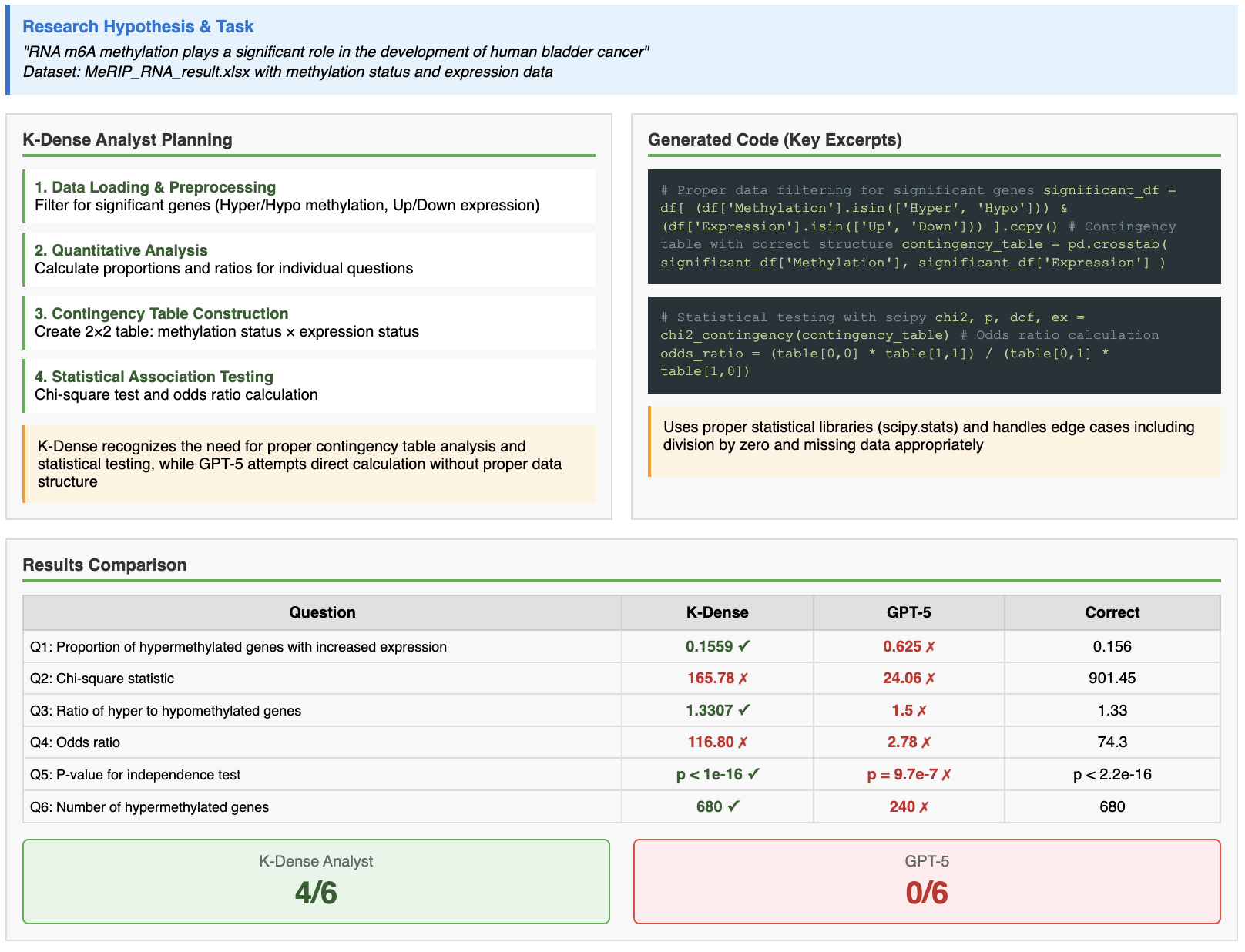}
  \caption{K-Dense Analyst's RNA Methylation Analysis (Bix-8). The dual-loop architecture enables systematic analysis of RNA m6A methylation in bladder cancer. The left panel shows the K-Dense Analyst's four-step planning process for data filtering, quantitative analysis, contingency table construction, and statistical testing. The right panel displays key code excerpts demonstrating proper use of pandas for data manipulation and scipy.stats for chi-square testing. The results table compares performance on six analytical questions, with K-Dense Analyst achieving 4/6 correct answers versus GPT-5's complete failure (0/6). The insight boxes highlight how K-Dense correctly implements contingency table analysis while GPT-5 attempts direct calculations without a proper data structure.}
  \label{fig:rna-methylation}
\end{figure}

\begin{figure}[h!]
  \centering
  \includegraphics[width=\textwidth]{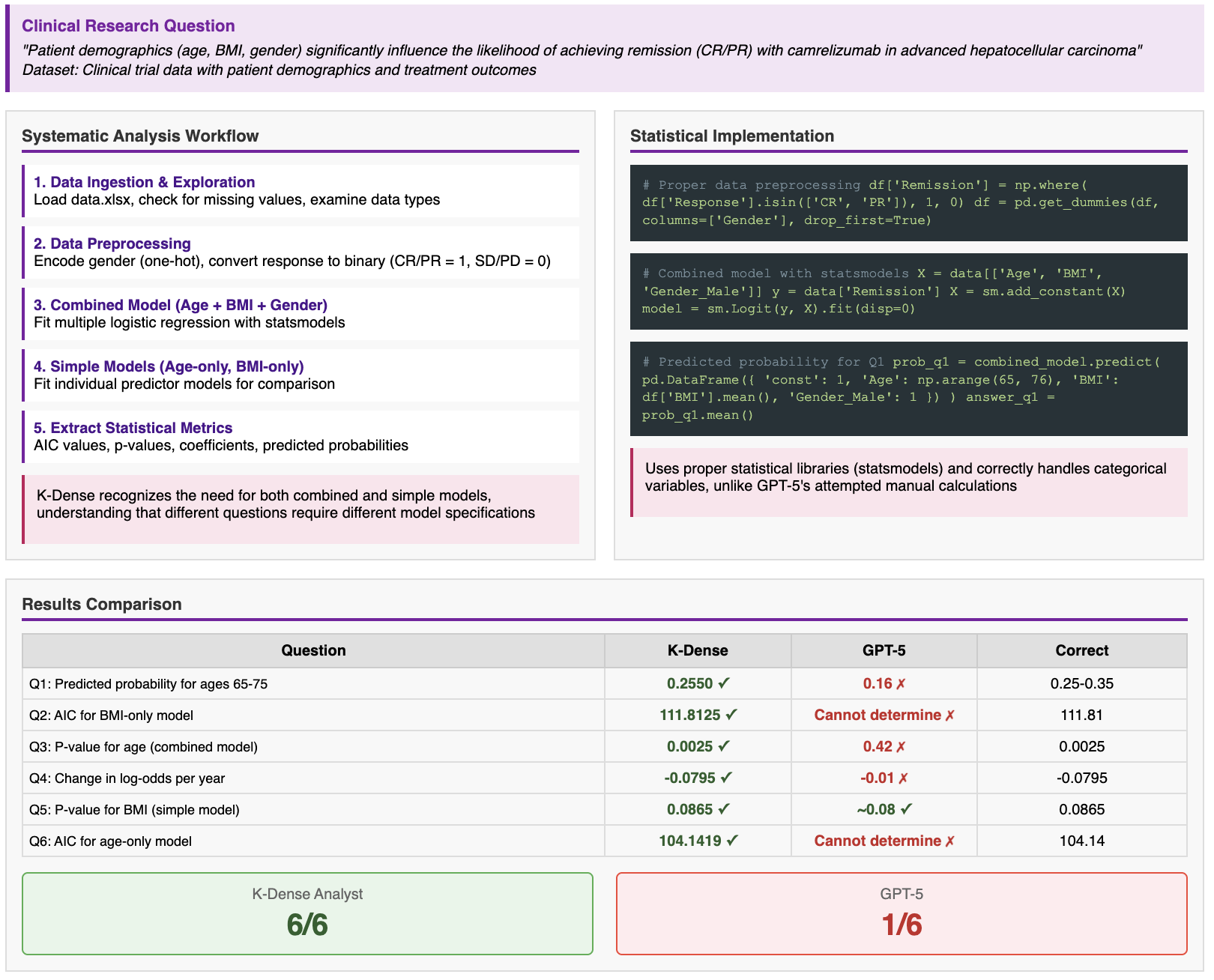}
  \caption{K-Dense Analyst's Logistic Regression Mastery (Bix-51). Demonstration of sophisticated statistical modeling capabilities on clinical trial data for camrelizumab treatment response. The left panel illustrates the systematic five-step workflow from data ingestion through model fitting to metric extraction. The right panel shows implementation using statsmodels for proper logistic regression with categorical variable handling. K-Dense Analyst achieves perfect accuracy (6/6) by correctly implementing both combined and simple models, extracting AIC values, and calculating predicted probabilities. In contrast, GPT-5 fails to specify models correctly and cannot extract key statistical metrics, achieving only 1/6 accuracy.}
  \label{fig:logistic-regression}
\end{figure}

\begin{figure}[h!]
  \centering
  \includegraphics[width=\textwidth]{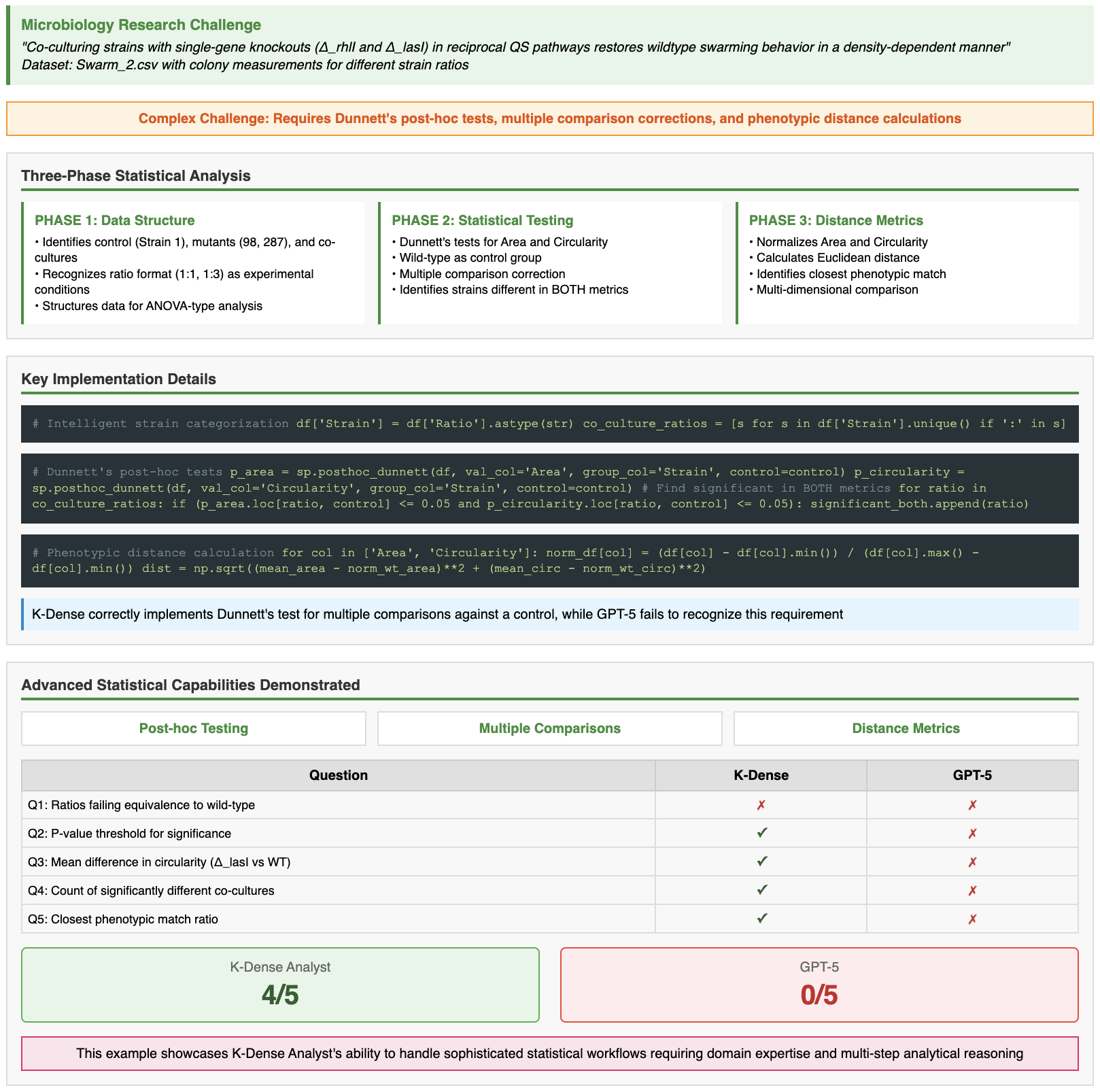}
  \caption{K-Dense Analyst Mastering Complex Multi-Comparison Testing (Bix-41). Analysis of microbial co-culture swarming behavior requires advanced statistical techniques. The three-phase approach progresses from data structure recognition through Dunnett's post-hoc testing to phenotypic distance calculations. Code snippets demonstrate proper implementation of multiple comparison corrections using scikit-posthocs and normalized Euclidean distance metrics. K-Dense Analyst successfully handles 4/5 questions requiring domain expertise in post-hoc testing, while GPT-5 fails completely (0/5), unable to recognize the need for Dunnett's test or implement proper multiple comparison corrections. The bottom panel emphasizes that this level of sophisticated statistical workflow is entirely beyond the capabilities of language-only models.}
  \label{fig:multi-comparison}
\end{figure}

These performance gains become more concrete when examining how K-Dense Analyst approaches specific challenges. Figure \ref{fig:rna-methylation} illustrates the system analyzing RNA methylation data, where it correctly implements chi-square tests and odds ratio calculations that GPT-5 fails to execute properly. The dual-loop architecture enables K-Dense Analyst to first plan the statistical approach, then generate validated code that correctly handles the contingency table construction.

Figure \ref{fig:logistic-regression} demonstrates even more sophisticated capabilities in logistic regression modeling. While GPT-5 struggles with basic model specification, K-Dense Analyst systematically builds multiple models, extracts AIC values, and correctly interprets p-values---achieving perfect accuracy across all six questions. This example highlights how the Coding Planning Agent's decomposition of complex statistical workflows enables accurate implementation.

Perhaps most impressively, Figure \ref{fig:multi-comparison} shows K-Dense Analyst handling Dunnett's post-hoc tests for co-culture experiments, a task requiring nuanced understanding of multiple comparison corrections and phenotypic distance calculations. The system's ability to normalize metrics and calculate Euclidean distances for phenotype matching demonstrates capabilities that extend far beyond pattern matching to actual scientific computation.

\section{Discussion}

The substantial performance improvements achieved by K-Dense Analyst highlight the value of specialized architectures for autonomous scientific reasoning in computational biology. Our results demonstrate that achieving meaningful autonomy in scientific analysis not only requires scaling language models but also demands purpose-built systems that can bridge the gap between high-level scientific questions and low-level computational execution.

The 34.4\% accuracy achieved by K-Dense Analyst on BixBench represents more than a benchmark improvement; it demonstrates that autonomous systems can now handle analytical challenges previously requiring human expertise. Tasks that stumped frontier language models, such as multi-step statistical analysis with appropriate corrections, integration of multiple omics datasets, or pathway enrichment with proper background selection, become tractable when approached through our architecture.

These performance gains merit closer examination of what we're actually comparing. K-Dense Analyst and the evaluated language models represent fundamentally different approaches to scientific analysis. Our system is an agentic framework with autonomous code execution, sandboxed tool use, and multi-agent validation loops. These capabilities enable it to solve entire categories of tasks requiring real computation, data transformation, or dynamic workflow adaptation. The fact that K-Dense Analyst achieves 34.4\% accuracy using Gemini 2.5 Pro as its base model, while Gemini 2.5 Pro alone achieves only 18.3\%, demonstrates that these architectural innovations unlock latent capabilities that cannot emerge from language modeling alone. Rather than viewing this as an unfair comparison, we should recognize it as proof that agentic designs are necessary for meaningful autonomous scientific analysis. Future benchmarking efforts should acknowledge these fundamental capability differences, potentially establishing separate evaluation tracks to better understand the contributions of different architectural approaches.

First, although BixBench currently stands as the most comprehensive benchmark for evaluating open-ended performance in bioinformatics agents, it is not without flaws. Through manual inspection, we found that at least one task, Bix10, contains an incorrect ground-truth answer. This discovery suggests that other tasks may also suffer from labeling errors or ambiguous evaluation criteria. Such inaccuracies undermine the validity of performance comparisons and may obscure meaningful model behavior. In fact, some apparent "mistakes" made by K-Dense Analyst may reflect shortcomings in the benchmark rather than factual analytical errors. For BixBench to serve as a reliable gold standard, the community must engage in systematic auditing and potentially develop a verified subset with expert-labeled annotations. Without such efforts, benchmarking results may overestimate or underestimate true system capabilities.

Second, while K-Dense Analyst achieves state-of-the-art results by leveraging frontier models like Gemini 2.5 Pro, these closed-source APIs limit reproducibility and accessibility for the broader research community. Fortunately, recent open-weight foundation models such as Qwen3 and DeepSeek R1 have shown competitive performance on science-centric tasks. We anticipate that replacing Gemini with Qwen3 will result in a modest drop in performance on BixBench, while significantly reducing costs and enabling full control over model deployment in both cloud and local settings. This suggests that the architectural contributions of K-Dense Analyst can be effectively reproduced using open models, and we plan to release configurations supporting this in the future. Expanding model support will help ensure that K-Dense Analyst can serve as a platform for inclusive and transparent experimentation.

Beyond model choices and benchmark quality, a third consideration is how systems handle the dynamic nature of scientific knowledge. K-Dense Analyst, while achieving state-of-the-art performance, encountered failures in tasks requiring very recent literature, newly introduced biological tools, or subtle parameter tuning based on emerging best practices. These limitations motivated the development of the broader K-Dense system, of which K-Dense Analyst is one component. The full K-Dense platform addresses these gaps through additional modules: a Tool Creation Agent that dynamically generates new computational tools on-demand, Deep Research capabilities providing real-time access to scientific databases and publications, and continuous knowledge integration mechanisms. These components work synergistically with K-Dense Analyst, enabling the complete system to handle the full research lifecycle from hypothesis generation through manuscript preparation. The system can recognize when existing tools are insufficient and create new ones, access the latest literature to inform analytical decisions, and even question potentially outdated benchmark answers when discrepancies are detected.

With these expanded capabilities, the complete K-Dense system is positioned to tackle even more challenging benchmarks. We plan to evaluate it against Humanity's Last Exam (HLE), which tests AI systems at the frontiers of human scientific knowledge across disciplines from physics to biology. Preliminary assessments suggest that the combination of K-Dense Analyst's analytical rigor with the broader system's research and tool creation capabilities could achieve competitive performance on this exceptionally challenging benchmark. Such results would demonstrate not just analytical competency but true scientific reasoning across domains.

The dual-loop architecture pioneered in K-Dense Analyst, which has proven effective in bioinformatics, extends naturally to other scientific domains. The fundamental principle, separating strategic planning from tactical execution with mandatory validation at both levels, addresses challenges common to all scientific analysis. In chemistry, the same architecture could orchestrate quantum calculations and molecular dynamics simulations. In climate science, it could manage complex Earth system models with appropriate uncertainty quantification. The key insight is that scientific analysis, regardless of domain, benefits from hierarchical decomposition with rigorous validation. We anticipate that domain-specific instances of this architecture, equipped with appropriate tools and validation logic, could accelerate discovery across the sciences.

Finally, deploying autonomous scientific agents raises deeper concerns around trust, transparency, and ethical oversight. While dual-agent cross-checking reduces the likelihood of hallucinated outputs, it does not eliminate the risk of subtle biases or failure modes that may arise in opaque model internals. In clinical or regulatory contexts, stronger auditability and standardization will be essential. The community should prioritize building agents with clear provenance trails, reproducible actions, and mechanisms for human override, especially when outputs inform high-impact decisions.

In light of these observations, several concrete next steps emerge. K-Dense Analyst and the broader K-Dense system will remain proprietary products, but we plan to publish detailed architectural specifications that others can use to build similar systems. We are developing API access to K-Dense Analyst for researchers who want to benchmark against it or use it for specific analyses. We'll contribute identified and corrected labels as well as documented issues back to the BixBench maintainers, improving benchmark reliability and benefiting everyone. We're also working on evaluation protocols that better distinguish between agentic and non-agentic systems, which we'll contribute to ongoing benchmarking discussions. On the technical development side, we're exploring continual knowledge ingestion and self-evaluation capabilities inspired by systems like STELLA, prioritizing wet-lab validation of our analytical outputs, and developing domain-specific versions for proteomics and structural biology. For the complete K-Dense system, we're exploring commercial partnerships where it makes sense. These steps reflect both the commercial realities of developing advanced AI systems and the genuine scientific value of the architectural patterns we've demonstrated. The dual-loop architecture and multi-agent validation approach can be implemented by others, even if our specific implementation remains proprietary.

\section{Conclusion and Future Work}

K-Dense Analyst advances the state of autonomous scientific reasoning by combining dual-loop planning, specialized validation, and modular tool integration to achieve state-of-the-art results on BixBench. Its 34.4\% accuracy represents not just an incremental improvement but a fundamental advance in how AI systems approach scientific analysis. By separating strategic planning from tactical execution and enforcing rigorous validation at every step, K-Dense Analyst demonstrates that autonomous scientific analysis is not only possible but practical.

The path forward is clear: autonomous systems must continuously integrate up-to-date scientific knowledge, adapt to emerging tools, evolve their own capabilities, and operate transparently with reproducible outputs. Our immediate goals include open-model deployments for accessibility, expanded data modality support, continual knowledge ingestion, and the development of self-evolving tools capable of adapting to workflows and methods without explicit definition. In parallel, we will contribute to improving benchmark quality and fairness, ensuring that future evaluations capture the full potential of agentic systems.

By refining architectures, broadening domain reach, enabling adaptive self-evolution, and deepening human-AI collaboration, we move closer to robust, trustworthy AI co-scientists capable of accelerating discovery across the life sciences and beyond.

\section{Methods}

We evaluate K-Dense Analyst using BixBench 1.0, a comprehensive benchmark developed by Mitchener et al. 2025 specifically designed to assess the capabilities of LLM-based agents in computational biology. BixBench represents a significant advancement in the evaluation of autonomous scientific systems, providing real-world analytical scenarios that capture the complexity and nuance of modern bioinformatics research.

BixBench comprises 53 expert-curated analytical scenarios ("capsules"), each representing authentic biological research questions drawn from published studies and real-world applications. These scenarios span diverse domains, including genomics, transcriptomics, proteomics, and systems biology, ensuring comprehensive coverage of modern bioinformatics workflows. Each scenario includes heterogeneous input data files, guiding research questions, and detailed analytical objectives that mirror the complexity faced by practicing bioinformaticians.

The benchmark includes 296 open-answer questions distributed across the 53 analytical capsules, with each capsule containing 3-7 associated questions for an average of 5.6 questions per scenario. These questions are designed to evaluate multiple aspects of analytical competency, including data exploration capabilities, multi-step reasoning, statistical analysis interpretation, and biological contextualization of results. The open-answer format is particularly challenging as it requires agents to generate free-form responses rather than select from predefined options, demanding deeper understanding and reasoning capabilities.

BixBench employs a rigorous evaluation framework that assesses both correctness and analytical depth. The primary metric focuses on open-answer accuracy, measured against expert-generated gold standard responses. Additionally, BixBench includes an optional multiple-choice question (MCQ) evaluation mode that provides complementary assessment of agent performance under different constraint conditions.

Following the protocols established by Mitchener et al. 2025, we evaluate K-Dense Analyst across the open answers task. The models are benchmarked on each capsule, receiving the research hypothesis, the question text, and available text and image data as the input. Text and image data that fit within the model context length are provided directly to the baseline models as part of the API call. In contrast, for K-Dense Analyst, the data are saved to the sandbox environment it has access to for analysis. All models and LLM API calls use the highest available reasoning setting.

A separate LLM-as-a-judge system, utilizing Gemini-2.5-pro with reasoning, evaluates the models' output against the ground truth and categorizes each question as either "correct" or "incorrect". The evaluation process accounts for the inherent variability in bioinformatics analyses by incorporating multiple valid solution pathways and assessment criteria that consider methodological appropriateness, result accuracy, and numerical stochasticity. All evaluations strictly adhere to the benchmark protocols to ensure fair comparison with baseline systems and reproducibility of results.

\bibliographystyle{unsrt}
\bibliography{K-DenseAnalystTowardsFullyAutomatedScientificAnalysis}

\end{document}